# Demonstration of real-time event camera to collaborative robot communication

Laura Duarte[1*], Michele Polito[2*], Laura Gastaldi[2], Pedro Neto[1], Stefano Pastorelli[2]

[1] University of Coimbra, Coimbra, Portugal
[2] Politecnico di Torino, Torino, Italy
* These authors contributed equally to this work
Corresponding author: michele.polito@polito.it (Michele Polito)

**Abstract:** Real-time robot actuation is one of the main challenges to overcome in human-robot interaction. Most visual sensors are either too slow or their data are too complex to provide meaningful information and low latency input to a robotic system. Data output of an event camera is high-frequency and extremely lightweight, with only 8 bytes per event. To evaluate the hypothesis of using event cameras as data source for a real-time robotic system, the position of a waving hand is acquired from the event data and transmitted to a collaborative robot as a movement command. A total time delay of 110 ms was measured between the original movement and the robot movement, where much of the delay is caused by the robot dynamics.

## 1 Introduction

Human-robot interaction is a dominant research topic, especially due to the exponential growth of AI technology in recent years [1, 2]. One of the main challenges when deploying collaborative systems is guaranteeing a short response time from the robot in order to provide prompt assistance to a human operator or activate safety procedures [3]. An optimal response time is usually called a real-time response, which is assumed to exist when the human operator has the feeling of an instantaneous response. The end-to-end response latency of the system can be measured to quantitatively evaluate the accuracy of real-time.

Traditional vision sensors, like the widespread frame-based cameras, provide a large amount of information about the environment and are backed by powerful recognition algorithms [4, 5]. However, this comes at the cost of a low rate of data transfer and the need to allocate memory for data storage. Event cameras were designed to transmit information asynchronously, transmitting information at a much higher frequency rate. For example, the Dynamic and Active-pixel Vision Sensor (DAVIS) can capture events up to a maximum bandwidth of 12 million events per second [6].

Since their development, event cameras have enabled many applications requiring low latency, such as autonomous drone flight, high-speed image reconstruction and optical flow estimation [7]. The deployment of collaborative robots in conjunction with event cameras is not new, but previous research efforts have focused mostly on real-time grasping strategies, which use an eye-in-hand configuration for the camera [8].

In this study, we propose tracking a free human arm gesture during handling in manufacture. A "hand waving" motion using a stationary DAVIS 240C event camera (iniVation, Zurich, Switzerland) has been captured. The location of the hand, defined as the region of interest (ROI), is found from the event data. The center coordinates of the ROI are transmitted through a UDP/IP connection to a Universal Robot 3 (UR3) (Universal Robots A/S, Odense, Denmark) to actuate its robotic arm in real-time. To accurately evaluate the total latency of the system, two Opal (Opal™ APDM, USA) inertial measurement units (IMUs) are mounted on the moving hand (master) and on the robot (slave) and the signal delay between the two is determined. Fig. 1 shows an overview of the proposed framework.

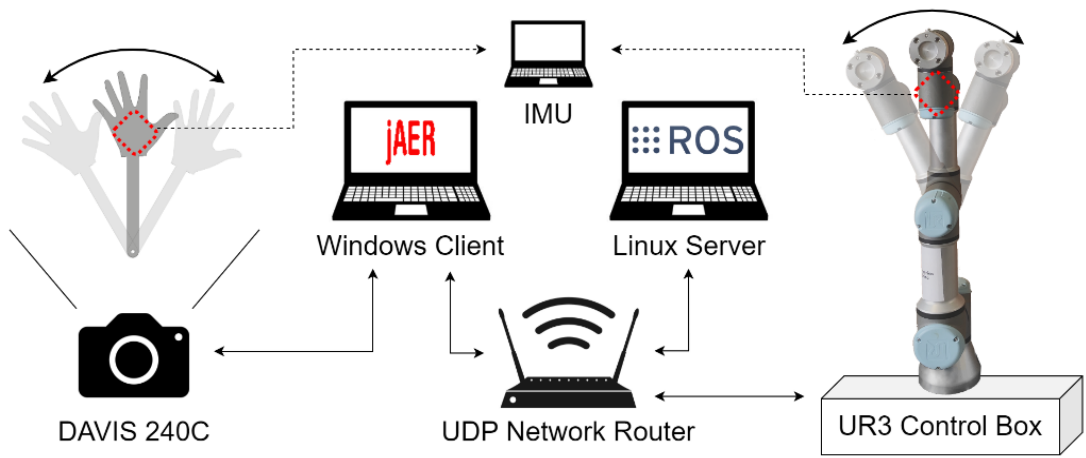

**Fig. 1.** Illustration of the proposed DAVIS 240C to UR3 framework.

## 2   Methodology

### 2.1   Hardware

The following experimental setup was deployed to use data obtained from an event camera to actuate the collaborative robot in real-time, as shown in Fig. 2. The main parts of the system are detailed as follows.

**Universal Robot 3 (UR3)**

The robot used in this work is the UR3, a 6-axes collaborative robot with a non-monocentric wrist. Each of its rotation axis is associated with a rotational joint, which can be identified as follows: Base, Shoulder, Elbow, Wrist 1, Wrist 2, and Wrist 3. In this work, only the Elbow rotational joint is actuated (Fig. 1). The UR3 robot has a maximum reach of 500 mm. The maximum tool center point linear speed is of 5000 mm/s. The first three joints can rotate at a maximum speed of 180 deg/s, while the three wrist joints can rotate twice as fast, at 360 deg/s. The maximum payload that the UR3 can support is 3 kg. The robot controller operates at a frequency of 125 Hz.

**Dynamic and Active Vision Sensor 240C (DAVIS240C)**

The event camera selected for the experimental setup is a Dynamic and Active-pixel Vision Sensor (DAVIS 240C). With a pixel array of 240 x 180 pixels, each of its pixels can output both absolute light intensity (synchronously, to create traditional frames) and changes in logarithmic light intensity (asynchronously, as events). In this work, only event data are captured and processed.

Each event $e$ generated by the DAVIS 240C is communicated as an array of values:

$$e = (x, y, ts, pol) \qquad (1)$$

where $x$ and $y$ represent the horizontal and vertical location of the event in the sensor array respectively, $ts$ is the timestamp of creation of the event with microsecond resolution, and $pol$ is the sign of the occurred brightness change.

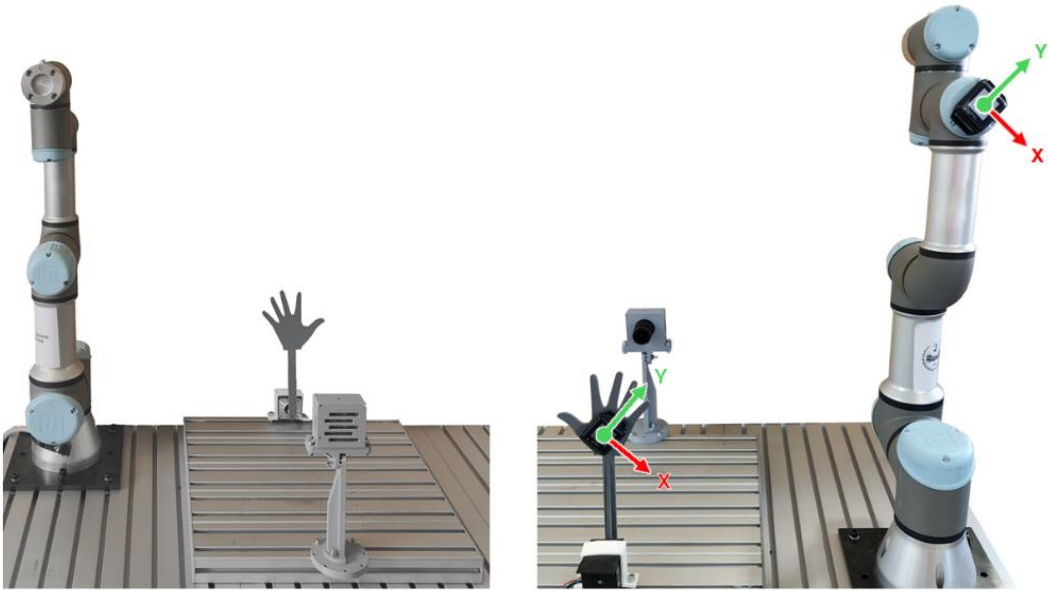

**Fig. 2.** Picture of the hardware setup featuring the UR3 robot, the DAVIS240C event camera and the two Opal IMUs.

**Opal Wearable Sensor**

The Opal is a small and lightweight wireless inertial measurement unit with a gyroscope, accelerometer, and a magnetometer. This wearable sensor was selected due to its high sampling rate, at 200 Hz, and accurate time synchronization, with precision of < 1 ms. To measure the latency of the system, one IMU wearable sensor is attached to the moving hand, and another is fixed to the robot arm. In this study only the gyroscope measurements were taken into account.

**Nema 17 Stepper Motor and Arduino Nano**

To actuate the hand waving, a Nema 17 stepper motor is used. Supplied with 24V through the TB6600 stepper driver, the stepper motor was set to rotate sinusoidally between [-36, 36°] at a frequency of 0.5 Hz. The motion frequency was selected based on two primary considerations: firstly, the frequency of human movements in industrial tasks is typically less than 1 Hz, and secondly, to ensure a safe distance from the robot's maximum dynamic capabilities and safety constraints. An Arduino Nano was used to program this movement.

## 2.2 Software

The software is split between two computers: a computer with a Windows Operating System (OS) serving as a client, to handle the event data, and another computer running Robot Operating System (ROS2 [9]) on a Linux OS serving as the server of the system, to receive this data and send movement commands to the robot.

### Windows Client

Event camera data are captured and visualized through the open-source software jAER [10]. This software is used to filter noise events and transmit the remaining events through a User Datagram Protocol (UDP) connection. Python code, running on the same computer, receives the data through this UDP connection. The events are then processed to find the location of the center of the ROI. The ROI's horizontal and vertical values range between [0, 240] pixels and [0, 180] pixels, respectively. To obtain the ROI, a method based on the detection of the object edge activity is deployed [11]. Resulting ROI center coordinates pass through a moving average filter, which averages the current value with the two previous ROI center coordinates values. This average is then finally transmitted through UDP communication to the second computer running ROS2.

### Linux Server

The program that manages communication with the camera is divided into two threads: one for UDP communication and another for publishing the received data from the event camera to the robot. The communication thread creates a UDP server that connects to the IP address of the Windows Client. Once communication is open, the thread waits for data to be sent from the client. As soon as data are received, they are processed. Specifically, the angle with respect to the vertical nominal axis of the ROI is calculated (Fig. 3a) to obtain the reference angle to be sent to the Elbow joint. An example of reference angle obtained from the experimental set up is shown in Fig. 3b. Finally, the processed data are stored in the global variable *msg*.

The second thread creates a ROS2 node, denominated *robot_reference*, that publishes a six-float array that contains the joints reference angle on the *forward_position_controllers/commands* topic. All the joints reference angles are constant except the Elbow one, that is equal to *msg*. Specifically, the node publishes the joints reference every 8 ms (125 Hz), regardless of whether the Elbow reference has been updated by the aforementioned thread. The graph of the ROS architecture is shown in Fig. 3c.

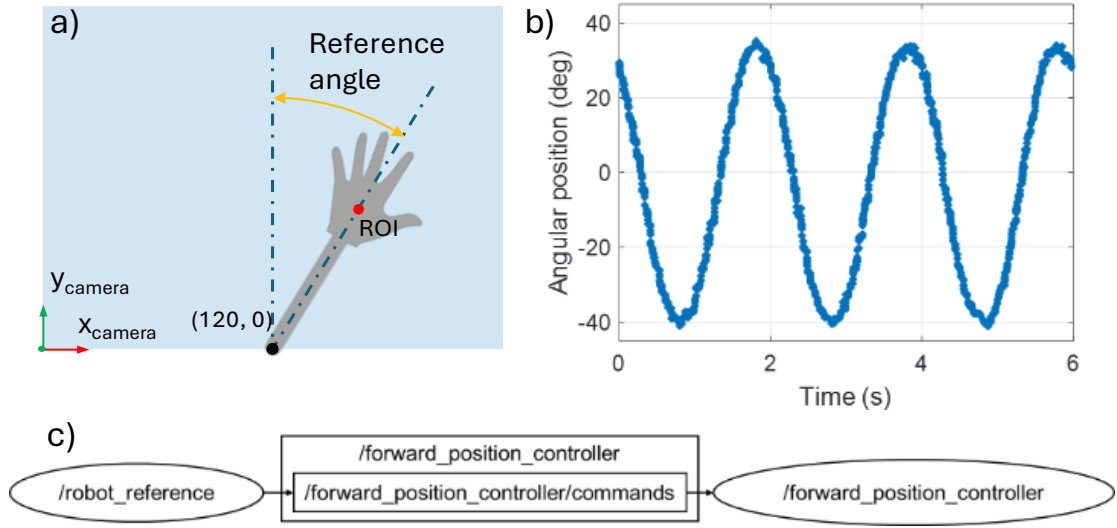

**Fig. 3.** a) Scheme of event camera frame and reference angle; b) Example of elbow joint reference angle obtained from the event camera; c) ROS graph.

Within the ROS2 Humble framework, the *forward_position_controller* from the Universal Robots ROS2 Driver package is used [12], which allows the robot to be commanded by specifying the desired position of each joint. It sends open-loop commands to the robot's internal controller, which is responsible for executing the command. Finally, the command is completed through a closed-loop control via an internal PID controller. The terms of the PID controller are tuned to optimize the dynamics of the robot in the required task. Specifically, it is possible to change the *gain* (ranging between 100-2000) and the *lookahead_time* (ranging between 0.03-0.2) In this study a *gain* of 1000 and a *lookahead_time* of 0.08 were selected. To be noticed that controller values drastically influence both the dynamic response of the robot and the system latency.

## 3       Results and discussion

The total latency of the system is measured to evaluate the performance of the robotic system (Fig. 4). To identify the main contributors to the total delay, the system is divided into its components, which are individually tested for their latency (Table 1).

### 3.1    Total System Latency

Two IMUs gyroscope's data in their z-axes, aligned with revolute hand an elbow joints respectively, are overlayed to accurately measure the total latency of the system, with temporal precision of < 1 ms (Fig. 4). The irregularities in the stepper signal are caused by the internal mechanism of the stepper motor which operates in discrete steps rather than a continuous motion. Using cross-correlation between the data generated from each of the IMUs, it was possible to determine that the system latency is 110 ms.

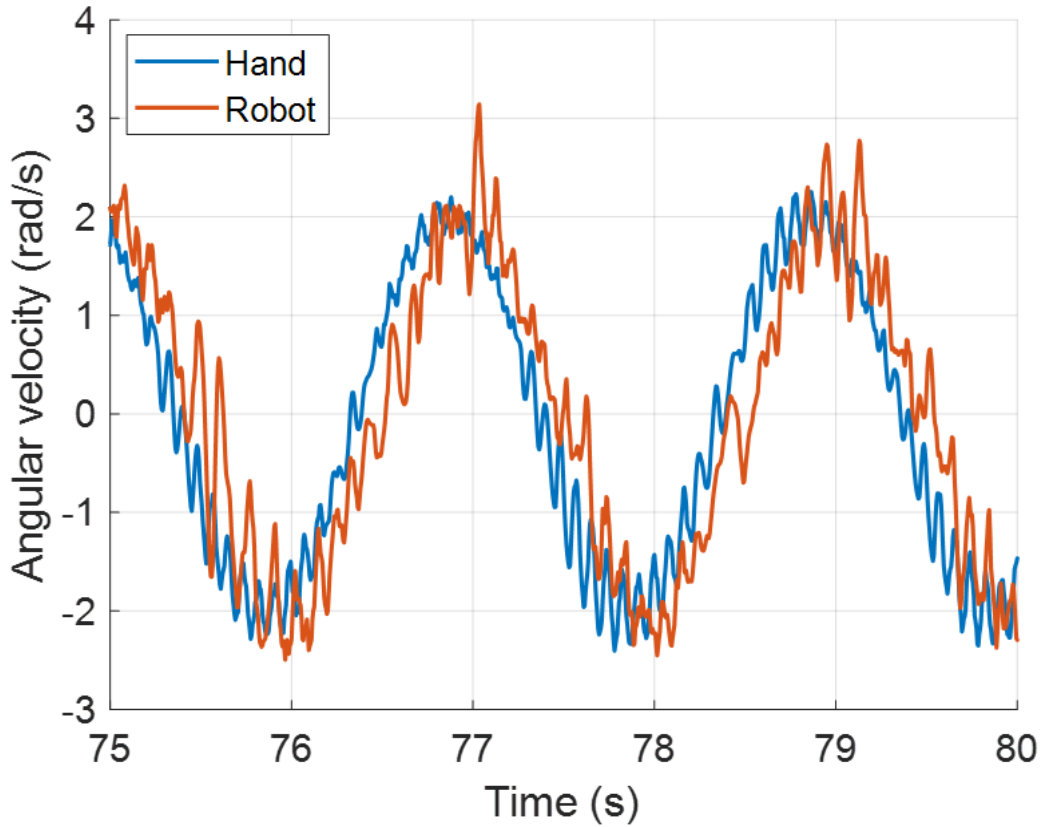

**Fig. 4.** Angular velocity registered on the moving hand compared to that of the robotic arm.

### 3.2 Components of System Latency

It is expected that the frequency of event generation will increase the faster the movement that is performed in front of the event camera. An event packet of 750 events is generated on average at 3.5 ms for a hand movement of 1 Hz and 6 ms for a movement at 0.5 Hz. Due to the inability to synchronize the event camera clock with the computer's system clock, there is no reliable way to measure the delay introduced in the system through the UDP communication. The Python program created to process the event data takes approximately 9 ms to retrieve encoded event data from the UDP connection, decode them, share them to a parallel process and compute the ROI.

After transmitting the data to the Linux computer, it needs to receive the data, run a function to transform the ROI center coordinates into the joint angle, and send this value to the robot control box. This section was evaluated with both a single-node and two-node ROS architecture, and the best results were obtained by the former, taking on average 4 ms to compute. Moreover, different trials were conducted to find the optimal *gain* and *lookahead_time* to balance robot responsiveness (minimizing latency) with robot behavior (minimizing instability).

**Table 1.** Components of system latency.

| Total latency of the system = 110 ms | | | | | | |
|---|---|---|---|---|---|---|
| Windows Client | | | | Linux Server | | UR3 Control Box |
| Capture events (750 events) | 1st UDP link | Event data processing | 2nd UDP link | Generate command | 3rd UDP link | Robot dynamics |
| ~ 6 ms | unknown | ~ 9 ms | ~ 4 ms | | | unknown |

**Conclusions**

Within the context of the proposed framework, the event camera proves to be a suitable sensor to provide data to actuate the robot in real-time, with a measured total system latency of 110 ms, which is considered real-time. Due to the sensor's high frequency rate, it can provide information at a frequency matching the robot controller. Furthermore, due to its direct relationship to movement in the environment, it will reduce or even stop dataflow when nothing is moving in the environment, thus not sending redundant and/or useless information to the system.

Most of the measured latency in the experiments takes place at the robot control box, so using a robot built for high-speed applications would significantly reduce the total latency of the system. A more complex application, with real human gestures, should be implemented to better understand the potential of the event camera for use in real-time human-robot collaboration scenarios.